\documentclass[preprint,12pt]{elsarticle}

\usepackage{amssymb}
\usepackage{booktabs} 
\usepackage{multirow} 
\usepackage{xcolor}   
\usepackage{float}
\usepackage{subfig}
\usepackage{algorithm}       
\usepackage{algpseudocode} 
\usepackage{todonotes}
\usepackage{url}
\usepackage{hyperref} 
\usepackage{tabularx}

\usepackage{todonotes}

\journal{Nuclear Physics B}

\usepackage{amsmath}

\begin{document}

\begin{frontmatter}

\title{Histo-MExNet: A Unified Framework for Real-World, Cross-Magnification, and Trustworthy Breast Cancer Histopathology}

\author[aff1]{Enam Ahmed Taufik}
\author[aff1]{Md Ahasanul Arafath}
\author[aff2]{Abhijit Kumar Ghosh}
\author[aff2]{Md. Tanzim Reza}
\author[aff3]{Md Ashad Alam}

\affiliation[aff1]{
    organization={European University of Bangladesh},
    department={Department of Computer Science and Engineering},
    city={Dhaka},
    country={Bangladesh}
}

\affiliation[aff2]{
    organization={BRAC University},
    department={Department of Computer Science and Engineering},
    city={Dhaka},
    country={Bangladesh}
}

\affiliation[aff3]{
    organization={Ochsner Center for Outcomes Research, Ochsner Research, Ochsner Clinic Foundation},
    city={New Orleans},
    state={LA},
    postcode={70121},
    country={USA}
}

\begin{abstract}
Accurate and reliable histopathological image classification is essential for breast cancer diagnosis. However, many deep learning models remain sensitive to magnification variability and lack interpretability. To address these challenges, we propose Histo-MExNet, a unified framework designed for scale-invariant and uncertainty-aware classification. The model integrates DenseNet, ConvNeXt, and EfficientNet backbones within a gated multi-expert architecture, incorporates a prototype learning module for example-driven interpretability, and applies physics-informed regularization to enforce morphology preservation and spatial coherence during feature learning. Monte Carlo Dropout is used to quantify predictive uncertainty. On the BreaKHis dataset, Histo-MExNet achieves 96.97\% accuracy under multi-magnification training and demonstrates improved generalization to unseen magnification levels compared to single-expert models, while uncertainty estimation helps identify out-of-distribution samples and reduce overconfident errors, supporting a balanced combination of accuracy, robustness, and interpretability for clinical decision support.
\end{abstract}


\end{frontmatter}



\section{Introduction}

Breast cancer remains a major cause of cancer-related mortality worldwide, making precise histological evaluation critical for reliable diagnosis and the planning of effective treatment strategies \cite{ref1, Yildiz2025}. Histopathology is still regarded as the gold standard for grading malignancy, as it provides rich information on tissue architecture, cellular morphology, and mitotic activity \cite{ref2}. In recent years, the introduction of Whole Slide Imaging (WSI) has transformed diagnostic workflows, moving pathology from conventional light microscopy towards digitally driven, computer-assisted analysis. This digitization has opened the door to the widespread use of deep learning methods, which seek to reduce inter-observer variability and enhance diagnostic consistency in routine practice \cite{ref3, ref4}. Within this landscape, convolutional neural networks have emerged as particularly powerful tools for feature extraction, achieving performance that approaches that of human experts in tasks such as subtype classification and mitosis detection \cite{ref5, ref6,alam2014kernel}.

Despite these algorithmic developments, there are still several major barriers to the therapeutic use of deep learning in pathology. First, pathologists use high magnification for cellular details and low magnification for architectural context, making histopathological diagnosis intrinsically multi-scale. Nevertheless, the majority of current deep learning models are trained on patches with a single magnification, which prevents them from generalizing over the "scale gap" and degrades their performance when used on unseen magnifications \cite{ref7}, \cite{ref8, Auliah-21}.  Second, a significant obstacle is data heterogeneity. Significant class imbalance and changes in staining procedures reported in real-world datasets lead to models that overfit to specific laboratory conditions rather than learning robust morphological features \cite{ref9}, \cite{ref11,AFROZ2024138,Hassan-21,Auliah-21}.

Most importantly, typical CNNs' "black-box" character erodes therapeutic trust.  When faced with out-of-distribution artifacts, conventional models frequently assign high confidence to wrong predictions (overconfidence) and produce a deterministic softmax probability \cite{ref13}.  Clinicians are unable to determine whether a model's choice is based on confounding noise or legitimate biological signals because of this lack of reliability.  Few studies have incorporated uncertainty quantification or physics-informed restrictions directly into the training aim to enforce biological plausibility, despite current research on explainable AI \cite{ref14, Alam-19}.

To address these challenges, this study introduces Histo-MExNet, a unified framework for scale-invariant, uncertainty-aware classification. The proposed framework is motivated by the practical limitations observed in real-world datasets, where class imbalance and substantial morphological heterogeneity often restrict the effectiveness of contemporary convolutional backbones. In addition, variations in image magnification introduce domain shifts that degrade feature consistency and limit generalization across unseen resolutions. Histo-MExNet is designed to overcome these constraints through a gated multi-expert ensemble trained on mixed magnification data, enabling the model to learn complementary feature representations while improving robustness and scale-invariant generalization. The architecture further incorporates a physics-informed regularization strategy to preserve morphological consistency during training, along with a prototype learning module that provides case-based explanations. By integrating uncertainty quantification with interpretable prototype representations, the proposed framework aims to reduce overconfidence and enhance clinical credibility without compromising predictive performance.

The main contributions of this work are summarized as follows:
\begin{itemize}
     \item We propose a gated multi-expert framework that integrates heterogeneous convolutional backbones to improve cross-magnification generalization and mitigate scale-dependent overfitting.
    
    \item We introduce a prototype learning module that structures class representations through multiple learnable exemplars, enabling example-driven interpretability and more transparent decision boundaries.
    
    \item We incorporate a physics-informed, domain-aware regularization strategy that constrains feature learning to preserve morphological coherence and reduce sensitivity to staining-related artifacts.
    
    \item We conduct comprehensive evaluations under same-magnification, cross-magnification, and mixed-magnification training protocols, demonstrating state-of-the-art performance on the BreaKHis dataset and improved robustness to magnification-induced domain shift.
 \end {itemize}

This paper's remaining sections are arranged as follows: Relevant literature on histopathological classification and uncertainty estimate is reviewed in Section \ref{sec:Literature review}.Section \ref{sec:Methodology} provides a detailed description of the proposed Histo-MExNet architecture, which includes the physics-informed loss and prototype module.  Section \ref{ER} presents the experimental setup, Ablation research, and quantitative results.  Finally, Section \ref{sec:discussion} discusses the clinical implications of our findings, and Section \ref{sec:conclusion} presents the findings.

\section{Literature review}
\label{sec:Literature review}
The transition from conventional manual microscopic inspection to digital pathology has revolutionized breast cancer diagnosis, enabling the application of deep learning methodologies for automated detection, subtyping, and prognostic forecasting. The initial emphasis of early research was on baseline classification with standard convolutional neural networks  \cite{hameed2022multiclass, ahmad_multires, li2020idsnet, jiang2019smallseresnet, zhu2019compactcnn, Alam-16a, Richfield-17}. This demonstrated that automated analysis could distinguish between tissue that was benign and tissue that was cancerous \cite{Paper04_AlexNet, Paper14_Araujo, Ashad-13,Ashad-08}.   A shift to increasingly complex tasks, like multi-class subtype classification, cross-magnification generalization, and clinically valid predictions, has been required due to the complexity of histopathology images. These problems have shown that typical CNNs have their limits and led to the creation of architectures that strike a compromise between accuracy, generalizability, and interpretabilit \cite{Alam-18C, Alam-21}.

The backbone design of a histopathology classifier is essential for extracting discriminative information from intricate tissue textures.  Early improvements depended a lot on making the network deeper to improve feature representation. Gour et al. \cite{Paper03_ResHist} demonstrated that a 152-layer ResHist model could overcome the issue of vanishing gradients and identify intricate semantic features at both cellular and tissue levels. Saini and Susan \cite{Paper07_GAN_Imbalance} also dealt with class imbalance, which is a common problem in histopathology datasets, by using VGG16 and GAN-based augmentation together. Although these methods emphasized the advantages of depth and augmentation, further examination indicated that deep networks alone were inadequate for capturing the intricate structural patterns essential for dependable categorization.

To solve this problem, DenseNet architectures, especially DenseNet201, have become the best way to get features from histopathology \cite{Paper16_DenTnet, Paper17_DenseGLCM}.  Dense connection improves the reuse of features, reduces the problem of disappearing gradients, and keeps both local and global tissue information. When used with textural descriptors like the Gray Level Co-Occurrence Matrix (GLCM), these networks are more sensitive to nuclear atypia and stromal patterns, which are critical for correctly telling the difference between subtypes. Recent studies showed that ConvNeXt and EfficientNet can achieve high predictive accuracy with reasonable computational cost \cite{xu_mdffnet, gurumoorthy_fsadl_bcdc}.  ConvNeXt can help to enhance the model's robustness to staining variability and other dataset-specific shifts \cite{NewPaper_ConvNeXt_2024}. In contrast, EfficientNetV2 uses dual squeeze-and-excitation modules to enhance parameter efficiency, but at the same time ensure strong performance on a range of classification tasks \cite{Paper20_DualSE, NewPaper_EfficientNet_Comparison}. These advances indicate the importance of an ensemble approach with different but complementary backbones. These combinations can improve how well things generalize and capture the multi-scale nature of histopathological images.

Pathologists can move seamlessly between scales to analyze microscopic cellular detail in routine practices. As a results a key difficulty emerged in computational pathology to understand the cellular details between scales. \cite{NewPaper_Magnification_2022}. Networks normally capture global context, but lose some morphological sign at low level of magnification. In contrast, it represents complex nuclear feature, but may miss broader architectural patterns in hight level of magnification. Sequence-based models such as Bi-LSTMs used to mitigate this issue in early works \cite{Paper05_BiLSTM}. Another early work showed that Capsule Networks can be used to preserve hierarchical and positional information \cite{Paper09_CapsNet}. Recent works suggest it shifted towards multi-scale ensemble methods that can combine predictions from several level of magnifications \cite{Paper22_Ensemble_WSI, Paper13_EnsembleWSI}. These studies show that combine output form different architecture with various scales it is possible to reduce the prediction variance of Whole Slide Image (WSI) classification \cite{liu_adaptivethresh}. This strategy closely parallels the way pathologists integrate localized detail with global context and provides a clear motivation for adopting a multi-expert ensemble design in the proposed Histo-MExNet framework.

 Although deep learning models are often accurate at classifying histopathological images, they are still not widely used in clinical workflows due to the lack of transparency in individual predictions \cite{gella_vit}. This becomes more difficult when there is a difference between the test slides and the training distribution \cite{yan_dwnatnet}. Uncertainty quantification has appeared as an promising strategy to improve the model's reliability in this context. Monte Carlo (MC) dropouts and deep ensemble methods can be used to estimate uncertainty and provide clinically meaningful confidence scores alongside predictions. \cite{NewPaper_UQ_BreastCancer_2023}. Interpret-ability remains essential to make things understandable. Prototype-based models associate input patches with learned class-specific examples  tend to offer more intuitive and clinically aligned explanations. Visually guides the decision by allowing experts to assess diagnostically relevant patterns in representative prototypes. Recent dual-prototype architectures \cite{NewPaper_DualPrototype_2024, NewPaper_PROTEGO_2025} further refine this idea by explicitly capturing both shared and class-specific characteristics that support more nuanced validation of model behavior against established pathological knowledge.

Conventional data driven models often tends to overlook the underlying biology of the tissue that can lead to exploit patterns that are not clinically meaningful. Physics-Informed Neural Networks (PINNs) and anatomically consistent learning frameworks incorporate domain knowledge directly into the training objective to mitigate this issue \cite{NewPaper_PINN_Review, NewPaper_AnatomicalConsistency_2024}. These methods have been used primarily in reconstruction and generative settings, but their application to classification problems is steadily increasing. Strategies that enforce morphological consistency by penalizing predictions that conflict with basic tissue structure that help guide models toward representations that better reflect both clinical expectations and histological reality.

These advances point to a broader shift from early CNN-based classifiers toward more integrated systems that can combine multiple expert backbones, handle uncertainty, and encode biological priors. Our proposed framework Histo-MExNet directly targets key challenges of robustness, generalization, and clinical trustworthiness by unifying diverse feature extractors, multi-scale representations, interpretable prototype mechanisms, and domain-aware regularization. This combination makes them especially promising for their real-world deployment in breast cancer histopathology.

\section{Methodology}
\label{sec:Methodology}

\subsection{Dataset Overview}

We used the Breast Cancer Histopathological Database (BreaKHis) in this study that consist 7,909 histopathological images of breast tumor tissue from 82 patients \cite{dataset}. The dataset contains 2,480 benign and 5,429 malignant samples that acquired at four level of optical magnifications (40$\times$, 100$\times$, 200$\times$, and 400$\times$) (Table~\ref{tab:distribution}). All tissue samples were obtained via surgical open biopsy and examined using immunohistochemistry. Sections of approximately 3~$\mu$m thickness were cut to preserve glandular and cellular architecture during preparation to make sure the resulting images are of sufficient quality for reliable computational analysis.

The availability of multiple magnification levels enables the dataset to capture both global tissue architecture and fine-grained cellular morphology. Lower magnifications (40$\times$, 100$\times$) show how the tissue is organized as a whole, whereas higher magnifications (200$\times$, 400$\times$) focus on distinctions at the cell level that are critical for telling tumor subtypes apart.  Table~\ref{tab:distribution} shows how many benign and cancerous samples there are at each level of magnification.

\begin{table}[H]
\centering
\caption{Summary of the BreaKHis dataset, detailing the number of benign and malignant histopathology images across four magnification levels. }

\label{tab:distribution}

\renewcommand{\arraystretch}{1.3}

\begin{tabular}{c c c c}
\hline
\textbf{Magnification Level} & \textbf{Benign} & \textbf{Malignant} & \textbf{Total} \\
\hline
40$\times$  & 625  & 1370 & 1995 \\
\hline
100$\times$ & 644  & 1437 & 2081 \\
\hline
200$\times$ & 623  & 1390 & 2013 \\
\hline
400$\times$ & 588  & 1232 & 1820 \\
\hline
\textbf{Total Images} & \textbf{2480} & \textbf{5429} & \textbf{7909} \\
\hline
\textbf{Total Patients} & \textbf{24} & \textbf{58} & \textbf{82} \\
\hline
\end{tabular}

\end{table}

\begin{figure}[H]
    \centering
    \includegraphics[width=0.9\linewidth]{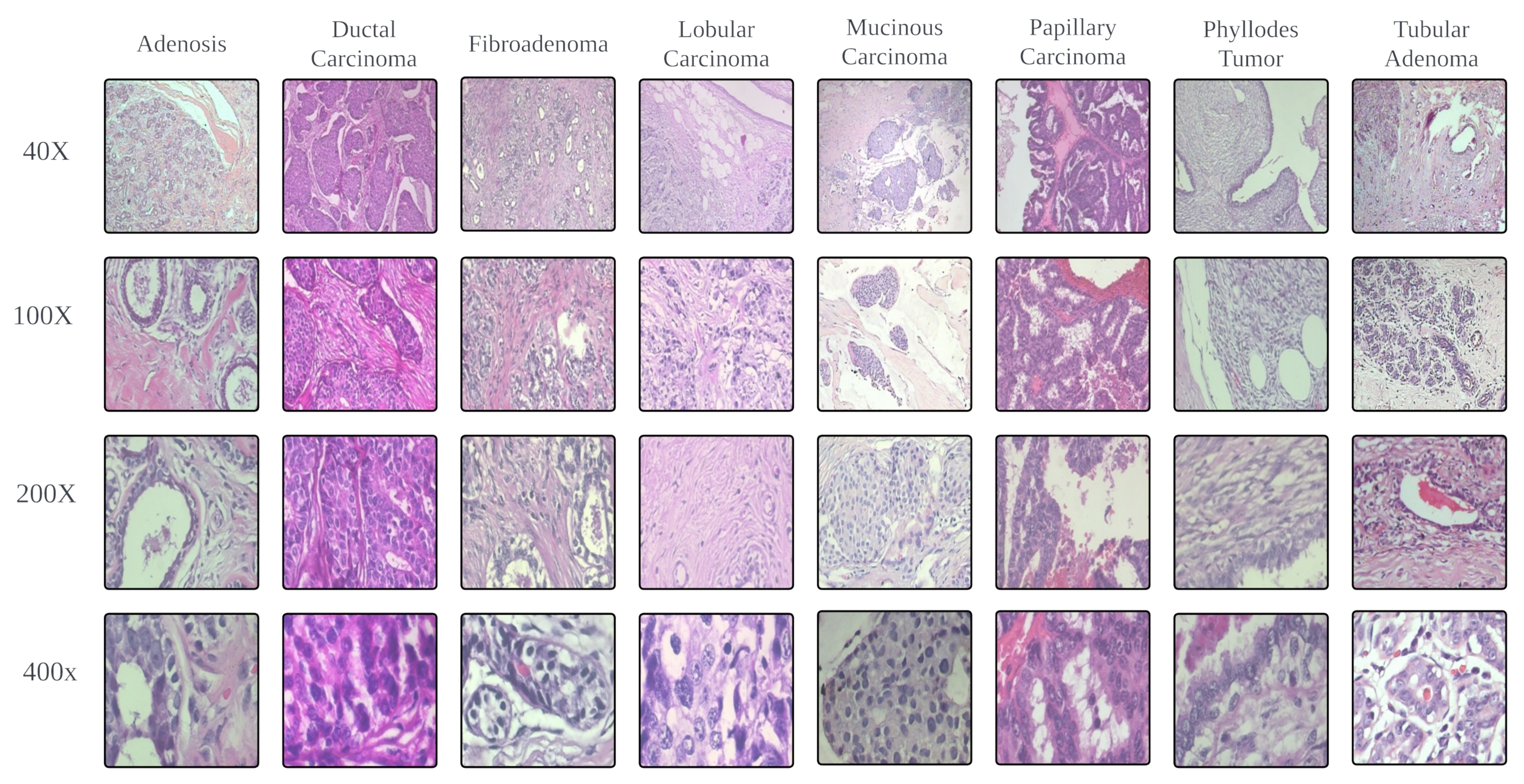}
    \caption{Representative subset of the BreaKHis dataset illustrating all eight tumor subtypes across four magnification levels ($40\times$, $100\times$, $200\times$, and $400\times$). Each column stands for a particular form of tumor, and each row stands for a different level of magnification. This shows how tissue morphology and cellular detail change at different scales.}

    \label{fig:dataset_subclass}
\end{figure}

Each image in the dataset belongs to one of eight histopathological subtypes (Fig.~\ref {fig:dataset_subclass}), divided into benign and malignant categories, as shown in Table~\ref{tab:subtypes}.

\begin{table}[H]
\centering
\caption{Overview of the benign and malignant histopathological subtypes represented in the BreaKHis dataset.}

\label{tab:subtypes}

\renewcommand{\arraystretch}{1.35}

\begin{tabularx}{\textwidth}{X X}
\hline
\textbf{Benign Tumors} & \textbf{Malignant Tumors} \\
\hline

\textbf{Adenosis}  
– Non-cancerous proliferation of glandular structures within the breast lobules. 
&
\textbf{Ductal Carcinoma}  
– Malignant tumor arising from epithelial cells of the milk ducts. \\
\hline

\textbf{Fibroadenoma} 
– Well-circumscribed benign mass composed of fibrous and glandular tissue.
&
\textbf{Lobular Carcinoma}  
– Invasive malignancy originating from the lobules, often with diffuse infiltration. \\
\hline

\textbf{Phyllodes Tumor}  
– Rare fibroepithelial lesion with variable behavior and potential for recurrence. 
&
\textbf{Mucinous Carcinoma}  
– Malignancy characterized by abundant extracellular mucin production. \\
\hline

\textbf{Tubular Adenoma}  
– Benign tumor consisting of tightly packed tubular structures with minimal atypia. 
&
\textbf{Papillary Carcinoma}  
– Malignant tumor exhibiting papillary architectural patterns. \\
\hline

\end{tabularx}

\end{table}

\subsection{Pre-processing}
All images in the BreaKHis dataset \cite{dataset} were processed through a systematic pre-processing pipeline to ensure data consistency and the reliability of model training. The dataset includes images at four magnification levels (40$\times$, 100$\times$, 200$\times$, and 400$\times$). All images were rescaled to a single size of $224 \times 224$ pixels to reduce the differences in spatial resolution. This pre-processing step standardizes spatial characteristics across magnifications that promotes consistent feature extraction with lower computational cost during training.

We had normalized all images to standardize the distribution of pixel intensities across the dataset after resizing. We had applied mean–variance (z-score) normalization to transform each pixel value, so that we have zero mean and unit variance. This  can also be written as:

\begin{equation}
I' = \frac{I - \mu}{\sigma},
\label{eq:normalization}
\end{equation}

here,

$I' =$ the normalized image,

$I =$ the original intensity of pixels,

$\mu = $ the mean intensity,

$\sigma =$ is the standard deviation computed over the dataset,

This normalization step helps to reduce the variability that is often found in histopathological slides caused by differences in illumination and staining. It also stabilizes the training process and supports efficient model convergence by providing a more consistent input distribution.

\subsection{Histo-MExNet Framework}

\begin{figure}[H]
    \centering
    \includegraphics[width=0.8\linewidth]{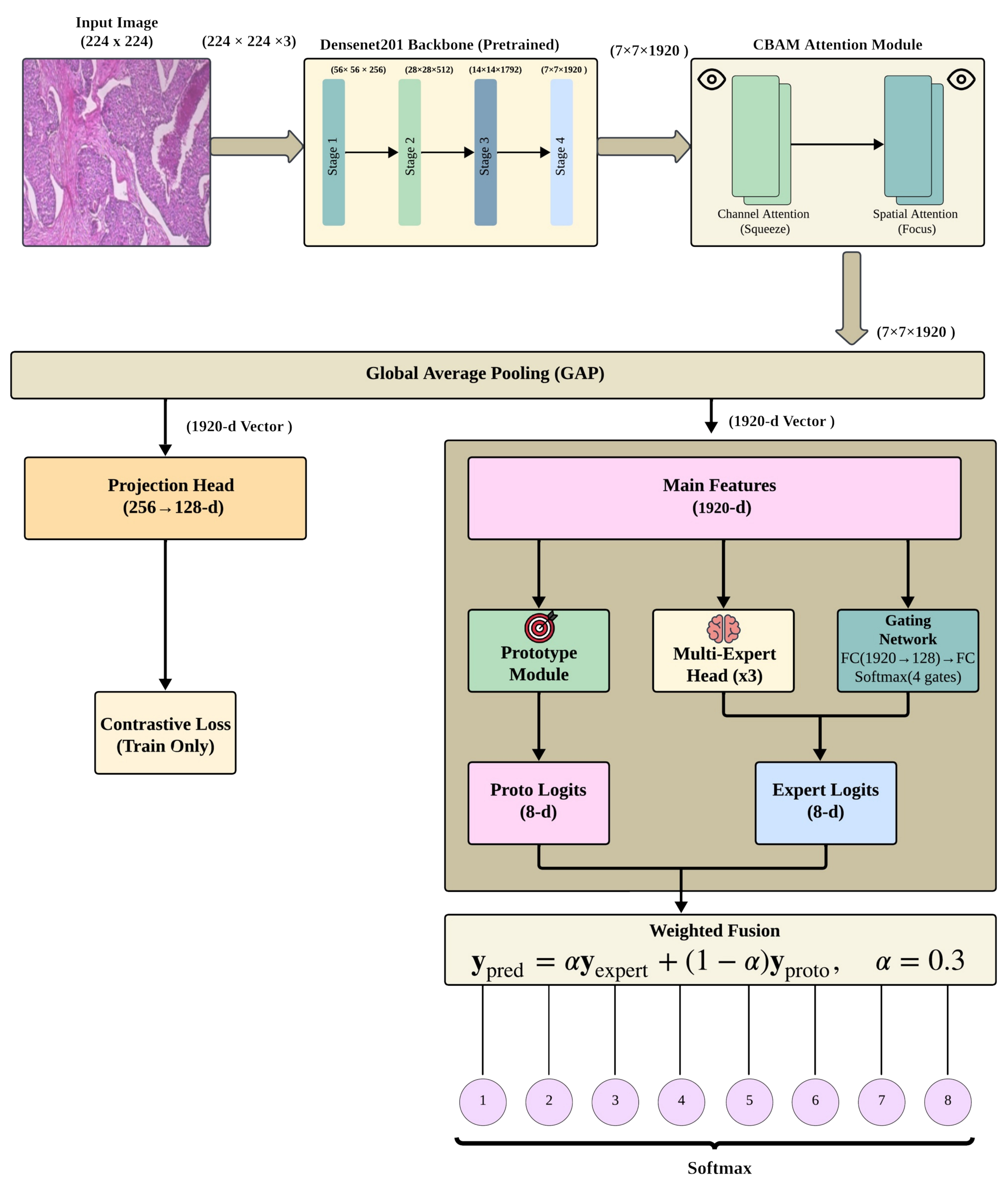}
    \caption{Overview of the proposed Histo-MExNet framework.Features are extracted using DenseNet201, the most effective backbone, as part with CBAM attention. These features are mapped to a prototype-aligned space and processed by several expert classifiers. The outputs of these classifiers are adaptively fused by a gating network to produce the final prediction.}
    \label{fig:Model_architecture}
\end{figure}

We propose Histo-MExNet shown in Fig.~\ref{fig:Model_architecture}, a unified framework designed for scale-invariant and uncertainty-aware classificationn. The model combines convolutional feature extractors with attention-based prototype-guided embeddings and a multi-expert ensemble to improve both predictive performance and reliability. Furthermore, the framework consists of physics-informed regularization with supervised contrastive learning and Bayesian uncertainty estimation to support robust behavior across heterogeneous tissue morphologies. Our Histo-MExNet framework provides uncertainty aware outputs alongside class predictions. It is designed to meet the needs of clinical practice and serve as a transparent decision support tool for breast cancer histology.

Histo-MExNet uses three complementary backbone architectures (DenseNet-201, ConvNeXt-Tiny, and EfficientNetV2-S) to extract multi-scale histopathological representations. 
Given an input image $\mathbf{x} \in \mathbb{R}^{3\times224\times224}$, backbone features are extracted, refined through attention mechanisms, and globally aggregated to obtain a unified representation:

\begin{equation}
\mathbf{f}_{\text{global}} = 
\mathrm{Concat}_k \Big(
\mathrm{GAP}\big(\tilde{\mathbf{F}}^{(k)}\big)
\Big)
\in \mathbb{R}^{3968},
\end{equation}

For supervised contrastive learning, $\mathbf{f}_{\text{global}}$ is further projected into a 128-dimensional embedding space $\mathbf{z}$ using a two-layer projection head.

To enhance classification robustness, we adopt a multi-expert learning paradigm with $K=3$ specialized expert heads $\mathcal{E}_k$ and one general classifier $\mathcal{G}$, each implemented with Monte Carlo Dropout. Their predictions are adaptively fused through a soft gating mechanism:

\begin{equation}
\mathbf{L}_{\text{expert}} = 
\sum_{k=1}^{K+1} g_k \, \mathcal{H}_k(\mathbf{f}_{\text{global}}),
\end{equation}

where $\mathcal{H}_k \in \{\mathcal{E}_1,\mathcal{E}_2,\mathcal{E}_3,\mathcal{G}\}$ and $g_k$ denote softmax-normalized gating weights.

To explicitly model class distributions, we introduce a prototype learning module in which each class $c$ is represented by multiple learnable prototypes $\{\mathbf{p}_{c,j}\}$. Prototype-based logits are computed via minimum normalized Euclidean distance:

\begin{equation}
\mathbf{L}_{\text{proto}}(c) 
= - \min_j d(\mathbf{f}_{\text{global}}, \mathbf{p}_{c,j}),
\end{equation}

Prototype parameters are optimized using a push--pull objective:

\begin{equation}
\mathcal{L}_{\text{proto}} =
D_{y} + 
\beta \max\left(0, \alpha + D_{y} - \min_{c\neq y} D_c \right),
\end{equation}

The final prediction is obtained through hybrid fusion of expert and prototype logits:

\begin{equation}
\mathbf{L}_{\text{final}} =
\lambda_1 \mathbf{L}_{\text{expert}}
+
\lambda_2 \mathbf{L}_{\text{proto}},
\end{equation}

To incorporate domain knowledge, physics-informed regularization is applied, including a biological plausibility constraint defined as:

\begin{equation}
\mathcal{L}_{\text{bio}} =
\|\mathbf{p} - \mathbf{R}\mathbf{p}\|_2^2,
\end{equation}

The overall optimization objective integrates focal loss, supervised contrastive learning, prototype learning, and physics-based regularization:

\begin{equation}
\mathcal{L}_{\text{total}} =
\sum_{i=1}^{6} \alpha_i \mathcal{L}_i,
\end{equation}

The overall operational flow of the proposed Histo-MExNet framework is summarized in Algorithm~\ref{alg:HistoMExNet}, outlining each central stage.

\begin{algorithm}[H]
\caption{Histo-MExNet: Multi-Expert Classification Framework}
\label{alg:HistoMExNet}
\begin{algorithmic}[1]
\State \textbf{Input:} Histopathological image dataset $\mathcal{D} = \{\mathbf{X}_i, y_i\}_{i=1}^N$
\State \textbf{Output:} Predicted class $\hat{y}_i$, confidence, and uncertainty for each image

\For{each image $\mathbf{X}_i \in \mathcal{D}$}
    \State Preprocess image: resize to $224 \times 224$ and normalize
    \For{each backbone $b \in \{\text{DenseNet201, ConvNeXt-Tiny, EfficientNetV2-S}\}$}
        \State Extract features: $\mathbf{F}_b = f_b(\mathbf{X}_i)$
        \State Apply attention: $\mathbf{F}'_b = \mathcal{A}(\mathbf{F}_b)$
        \State Project to prototype-aligned embedding: $\mathbf{z}_b = g(\mathbf{F}'_b)$
    \EndFor
    \For{each expert classifier $\mathcal{E}_k$}
        \State Compute class probabilities: $\mathbf{p}_k = \mathcal{E}_k(\mathbf{z}_b)$
    \EndFor
    \State Fuse expert predictions via gating network: $\hat{\mathbf{p}}_i = \sum_k \alpha_k \mathbf{p}_k$
    \State Compute physics-informed losses: 
    \[
        \mathcal{L}_{\text{morph}},\ \mathcal{L}_{\text{spatial}},\ \mathcal{L}_{\text{bio}}
    \]
    \State Compute total loss: $\mathcal{L}_{\text{total}} = \mathcal{L}_{\text{CE}} + \lambda_1 \mathcal{L}_{\text{morph}} + \lambda_2 \mathcal{L}_{\text{spatial}} + \lambda_3 \mathcal{L}_{\text{bio}}$
    \State Update network weights using $\mathcal{L}_{\text{total}}$
    \State Estimate uncertainty using Monte Carlo Dropout: $\text{uncertainty}(\mathbf{X}_i)$
    \State Determine predicted class and confidence: $\hat{y}_i = \arg\max \hat{\mathbf{p}}_i$, $\text{confidence} = \max \hat{\mathbf{p}}_i$
\EndFor

\State \textbf{return} $\{\hat{y}_i, \text{confidence}, \text{uncertainty}(\mathbf{X}_i)\}_{i=1}^N$
\end{algorithmic}
\end{algorithm}

The performance of our proposed Histo-MExNet framework was evaluated using standard classification metrics, along with uncertainty and confidence measures. Accuracy measures the proportion of correctly classified samples:
\begin{equation}
\text{Accuracy} = \frac{\text{Number of correctly classified samples}}{\text{Total number of samples}}.
\end{equation}

To address the issue of class imbalance, we report weighted Precision, Recall, and F1-Score, with each class weighted according to its relative frequency within the dataset. The reliability of the model was evaluated by measuring the predictive uncertainty and confidence. Average Uncertainty quantifies the mean variability in the predicted class probabilities over all samples. On the other hand, the average confidence corresponds to the average of the highest predicted probability for each sample. When both of these measures were combined, it gave a complete view of both the accuracy and the trustworthiness of the output of the model. We additionally report Correct Confidence and Wrong Confidence to further assess the calibration quality that can be defined as follows:
\begin{align}
\text{Correct Confidence} &= \frac{1}{|\mathcal{C}|} \sum_{i \in \mathcal{C}} \max(\mathbf{p}i),\\
\text{Wrong Confidence} &= \frac{1}{|\mathcal{W}|} \sum{i \in \mathcal{W}} \max(\mathbf{p}_i),
\end{align}
Here, 

$\mathcal{C} =$ sets of correctly classified samples,

$\mathcal{W} =$ sets of incorrectly classified samples,

$\mathbf{p}_i =$ the predicted probability vector for the $i^{th}$ sample.

\section{Experiments and Results}
\label{ER}

\subsection{Step-wise Evaluation of Histo-MExNet Generalization}
\label{subsec:first_stage}

We took our experimental analysis on a journey through three distinct stages. In the first stage, the model was trained and tested on the same magnification data using a stratified random split, and performance was evaluated at each magnification. The second stage involved cross-magnification evaluation, in which the model trained at one magnification (100x, selected for its balanced performance) was tested at other magnifications (40x, 200x, and 400x) to assess generalization. In the final stage, the model was trained on a combined dataset across all magnifications and evaluated on a stratified split to assess its robustness and scale-invariant learning.

In our first stage, we established a baseline by training several state-of-the-art models without augmentation or rebalancing to measure how different convolutional architectures perform on real-world imbalanced data. The result (Table \ref{tab:performance_baseline_models}) was not promising and showed high variance across the data, with a minimum accuracy of 45\% and a maximum of 82\%. It shows that naive transfer learning is ineffective due to variation in the dataset's physical form and structure. Based on these observations, three representative architectures were selected for subsequent experiments: Efficient Net V2-S, DenseNet201, and ConvNeXt-Tiny. DenseNet201 was selected for its deep, balanced representational capacity, Efficient NetV2-S for its lightweight yet dependable performance, and ConvNeXt-Tiny for its cutting-edge convolutional design inspired by ResNet architectures. These models collectively provide a broad yet complementary range of depth, efficiency, and modern design, laying the groundwork for our future experimental research.

\begin{table}[H] 
\centering
\caption{Comparison between baseline CNN architectures evaluated under Type 1 experimental settings, where models are trained and tested on the same magnification level using a random split on each type of magnification.}

\label{tab:performance_baseline_models}
\tnotetext[tn1]{Maximum (best) values in each column are highlighted in \textbf{bold}.}

\begin{tabular}{l l r r r r} 
\toprule
\textbf{Backbone Model} & \textbf{Metric} & \textbf{40X} & \textbf{100X} & \textbf{200X} & \textbf{400X} \\
\midrule
\multirow{2}{*}{VGG16} 
& Accuracy & \textbf{67\%} & 58\% & 52\%  & 52\%\\
\cmidrule(lr){2-6} 
& F1-Score & \textbf{0.59} & 0.42 & 0.26 & 0.37\\
\cmidrule(lr){1-6} 

\multirow{2}{*}{ResNet50} 
& Accuracy & 49\%  & 49\% & \textbf{53\%} & 47\% \\
\cmidrule(lr){2-6} 
& F1-Score & 0.36 & 0.32 & \textbf{0.38} & 0.33 \\
\cmidrule(lr){1-6}

\multirow{2}{*}{EfficientNetB0 \textcolor{red!80!black}{\large\textbf{\(\downarrow\)}}} 
& Accuracy & 48\% & \textbf{49\%} & 46\% & 45\% \\
\cmidrule(lr){2-6} 
& F1-Score & \textbf{0.23} & 0.22 & 0.20 & 0.19 \\
\cmidrule(lr){1-6}

\multirow{2}{*}{DenseNet201 \textcolor{green!50!black}{\large\textbf{\(\star\)}}} 
& Accuracy & 80\% & \textbf{82\%} & 76\% & 74\% \\
\cmidrule(lr){2-6} 
& F1-Score & \textbf{0.76} & 0.78 & 0.69 & 0.70 \\
\bottomrule
\end{tabular}
\end{table}

We added our proposed Histo-MExNet pipeline to the representative architectures we had chosen based on baseline performance to test how well it worked on real-world imbalanced data. In the first-stage experiments, it was found to have a minimum accuracy of 88.46\% and a maximum of 98.25\%, with a much higher weighted F1-Score at all magnifications (Table \ref{tab:hybrid_model_performance_merged}).

\begin{figure}[H]
    \centering
    \subfloat[t-SNE using traditional CNNs]{\includegraphics[width=0.48\linewidth]{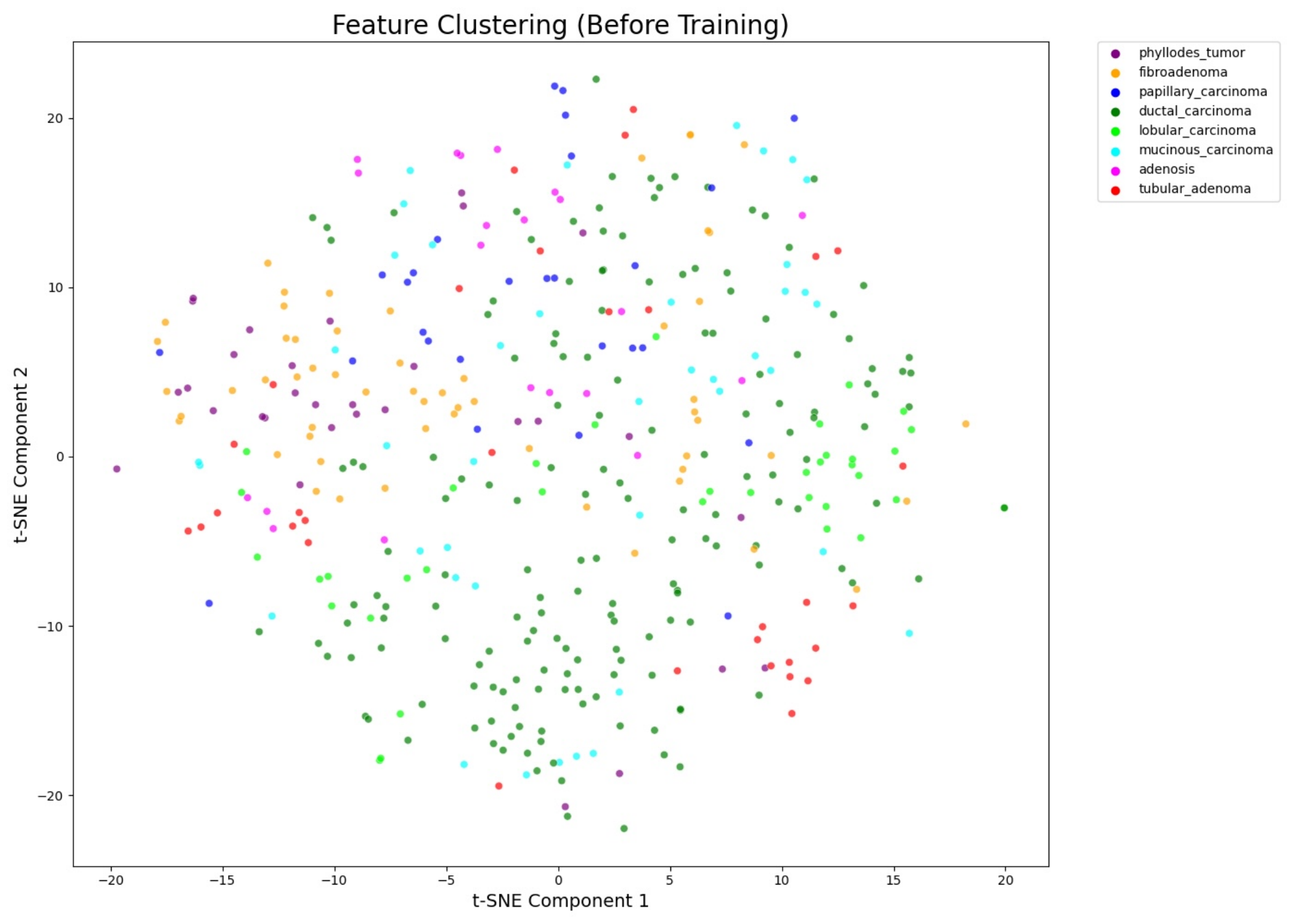}
    \label{fig:tsne_baseline_100X}} 
    \hfill
    \subfloat[t-SNE plot after training on type 1 experimental condition]{\includegraphics[width=0.48\linewidth]{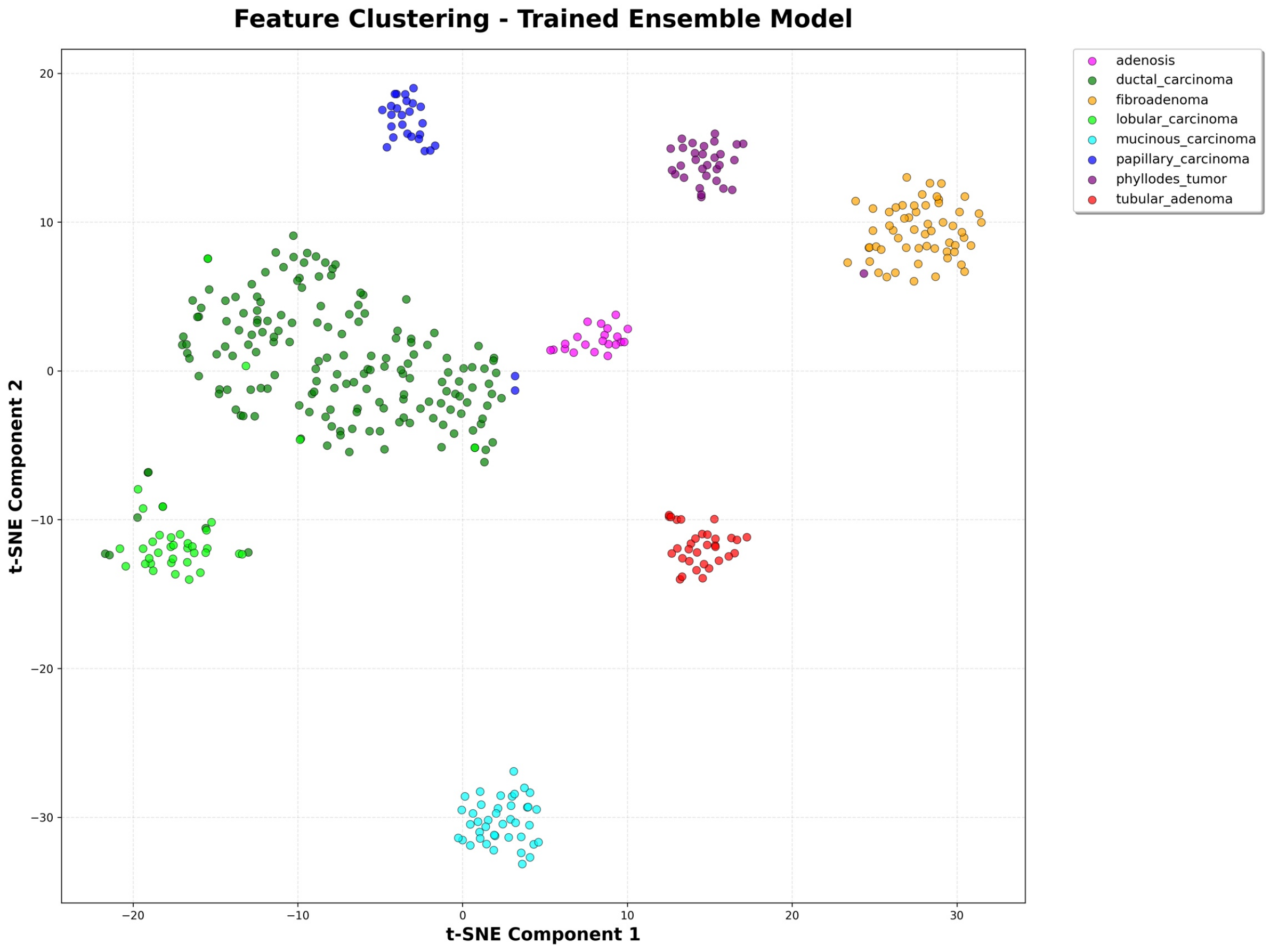}
    \label{fig:tsne_proposed_100X}}
    \caption{t-SNE visualization comparing how well the baseline CNNs and the proposed model using the DenseNet201 backbone separate features at a $100\times$ magnification.}

    \label{fig:tsne_analysis}
\end{figure}

Figure \ref{fig:tsne_analysis} shows that visual analysis backs up this improvement in performance even more. Fig. \ref{fig:tsne_baseline_100X} shows that the feature space of traditional CNN models has poorly separated clusters. All eight classes are mixed together in a single, chaotic cloud, which is why baseline models don't converge well. In contrast, our proposed 5-fold ensemble training produces a feature space that shows eight distinct, compact, and well-separated clusters for each class (Fig. \ref{fig:tsne_proposed_100X}). In the first stage, our hybrid model with a DenseNet201 backbone, hybrid loss, and 5-fold cross-validated ensemble learning does much better than baseline architectures in both classification accuracy and feature space separability.

When looked at with other levels of magnification in the second stage, it was clear that the model's performance decreased a lot as the magnification increased from 100x to 400x (Table \ref{tab:hybrid_model_performance_merged}). The proposed model did very well at 40x (about 78\%), but only about 51\% at 400x. All of the backbones behaved the same way, which highlight that the dataset's scale-specific morphological changes caused a shift in the domain.

\begin{table}[H] 
\centering
\caption{Performance of the proposed Histo-MExNet framework under Same-Magnification (Type 1) and Cross-Magnification (Type 2) experimental settings.}
\label{tab:hybrid_model_performance_merged}
\tnotetext[tn2]{Maximum (best) values in each column are highlighted in \textbf{bold};}

\resizebox{\textwidth}{!}{%
\renewcommand{\arraystretch}{1.2} 
\begin{tabular}{l | l | r r r r | r r r} 
\toprule
\multirow{2}{*}{\textbf{Backbone Model}} & \multirow{2}{*}{\textbf{Metric}} & \multicolumn{4}{c|}{\textbf{Type 1: Same-Magnification}} & \multicolumn{3}{c}{\textbf{Type 2: Cross-Magnification}} \\
\cmidrule{3-9}
& & \textbf{40X} & \textbf{100X} & \textbf{200X} & \textbf{400X} & \textbf{40X} & \textbf{200X} & \textbf{400X} \\
\midrule
\multirow{2}{*}{DenseNet201} 
& Accuracy & \textbf{98.25\%} & 96.88\% & 92.80\% & 88.46\% & \textbf{73.08\%} & 71.93\% & 51.15\% \\
& F1-Score & \textbf{0.983} & 0.969 & 0.927  & 0.881 & \textbf{0.716} & 0.690 & 0.393 \\
\hline 

\multirow{2}{*}{ConvNeXt-Tiny} 
& Accuracy & 95.74\% & \textbf{97.84\%} & 92.8\% & 90.66\% & \textbf{78.25}\% & 73.27\% & 51.37\% \\
& F1-Score & 0.956 & \textbf{0.978}  & 0.926 & 0.906 & \textbf{0.776} & 0.708  & 0.397 \\
\hline

\multirow{2}{*}{EfficientNetV2-S \textcolor{red!80!black}{\large\textbf{\(\downarrow\)}}} 
& Accuracy & \textbf{95.99\%} & 92.09\% & 93.55\% & 89.84\% & 67.92\% & \textbf{72.03\%} & 55.22\% \\
& F1-Score & \textbf{0.959} & 0.921 & 0.935 & 0.898 & 0.665 & \textbf{0.694} & 0.482\\
\bottomrule
\end{tabular}%
}
\end{table}

In the third stage, we trained our proposed model, Histo-MExNet, on a dataset that included all four magnifications. A custom key was made to make sure that the data split was consistent and representative, uniquely encoding both the class label and the level of magnification. With this stratified key, a 20\% split (1582 samples) was done to make sure that each magnification and class was equally represented. In this setup, all hybrid variants had accuracies of over 93\%, which shows that they were able to adapt to different scales (table \ref{tab:hybrid_model_performance_all_mag}). Joint training on the combined dataset improved both accuracy and reliability compared to earlier single-magnification experiments, and it also reduced biases that were specific to magnification. In this experiment, our proposed model with a DenseNet201 backbone performed better than the other two backbone architectures.

\begin{figure}[H]
    \centering
    \subfloat[ConvNeXt Uncertainty]{\includegraphics[width=0.23\linewidth]{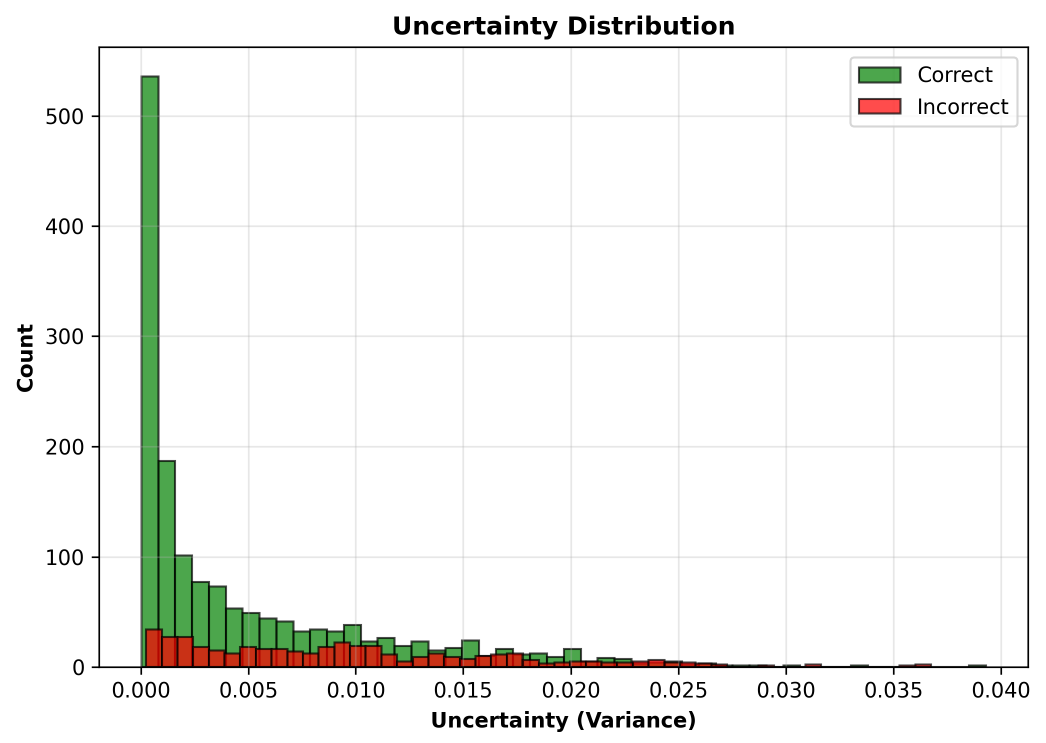}\label{fig:uncertainity_convNext}} 
    \hfill
    \subfloat[EfficientNet Uncertainty]{\includegraphics[width=0.23\linewidth]{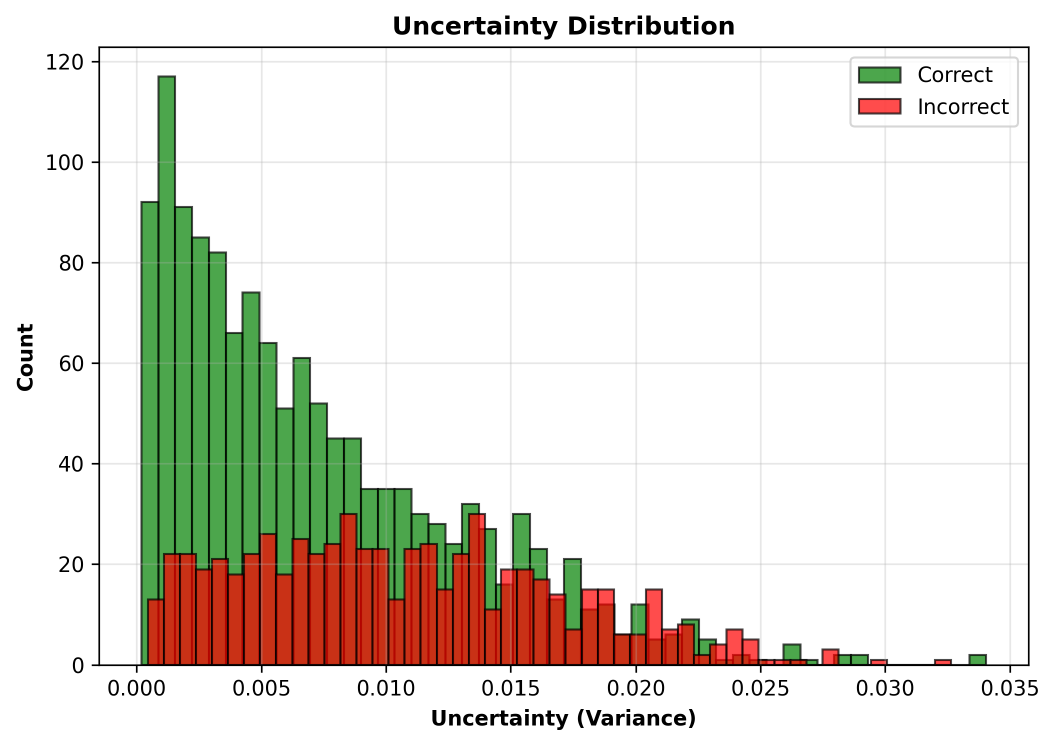}\label{fig:uncertainity_efficientnet}}
    \hfill
    \subfloat[ConvNeXt Confidence]{\includegraphics[width=0.23\linewidth]{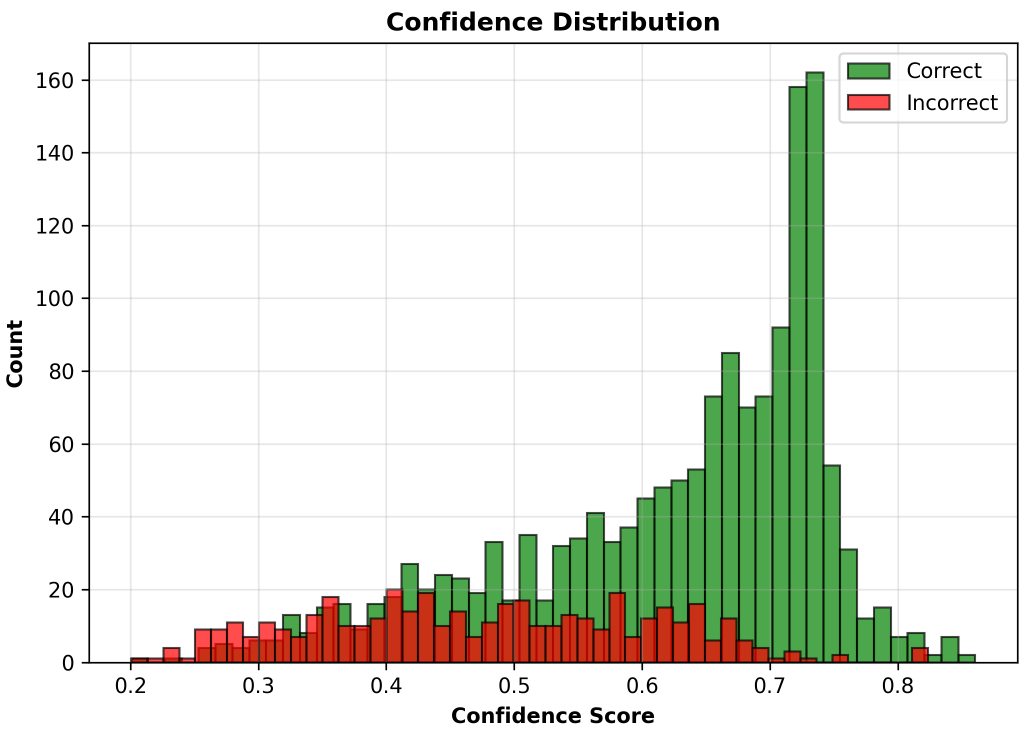}\label{fig:confidence_convnext_100x_40x}} 
    \hfill
    \subfloat[EfficientNet Confidence]{\includegraphics[width=0.23\linewidth]{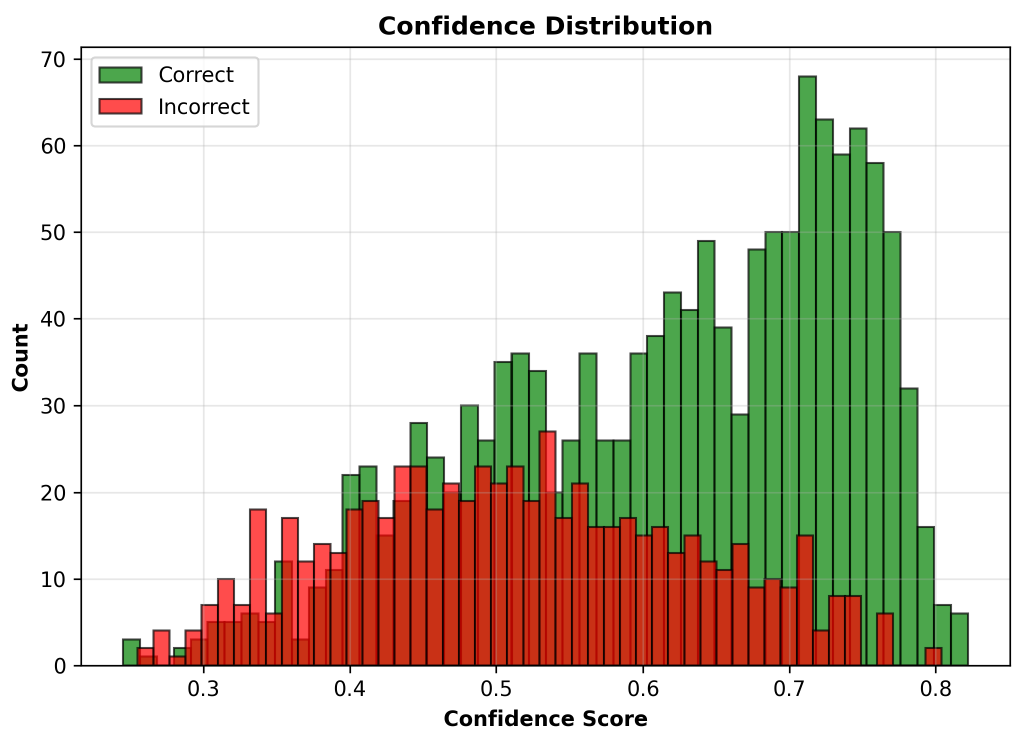}\label{fig:confidence_efficientnet_100x_40x}}
    
    \vspace{0.5em} 
    
    \subfloat[DenseNet201 Variance]{\includegraphics[width=0.23\linewidth]{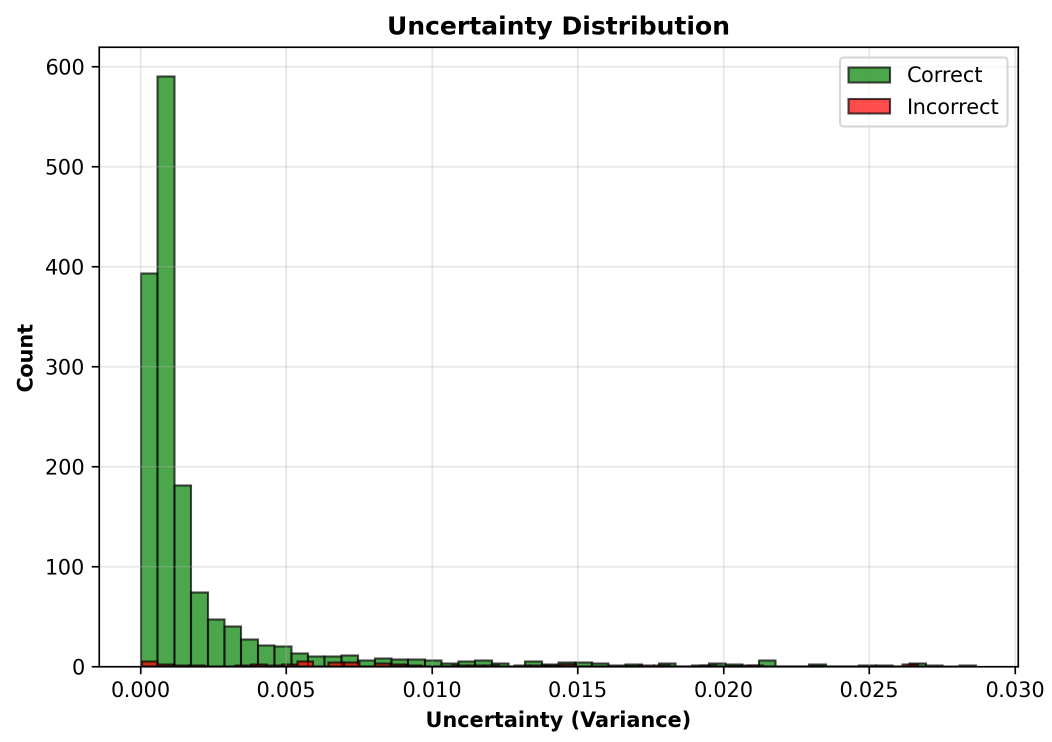}\label{fig:uncertainty_all_mag_densenet}} 
    \hfill
    \subfloat[ConvNeXt Variance]{\includegraphics[width=0.23\linewidth]{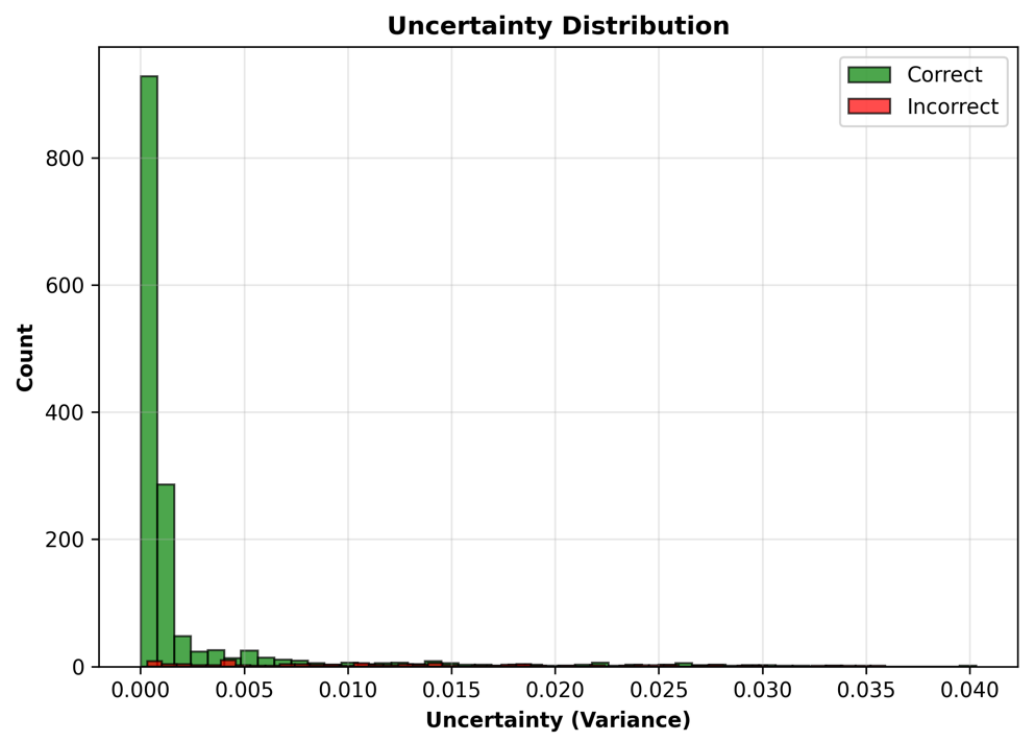}\label{fig:uncertainty_all_mag_convnext}}
    \hfill
    \subfloat[DenseNet201 U vs C]{\includegraphics[width=0.23\linewidth]{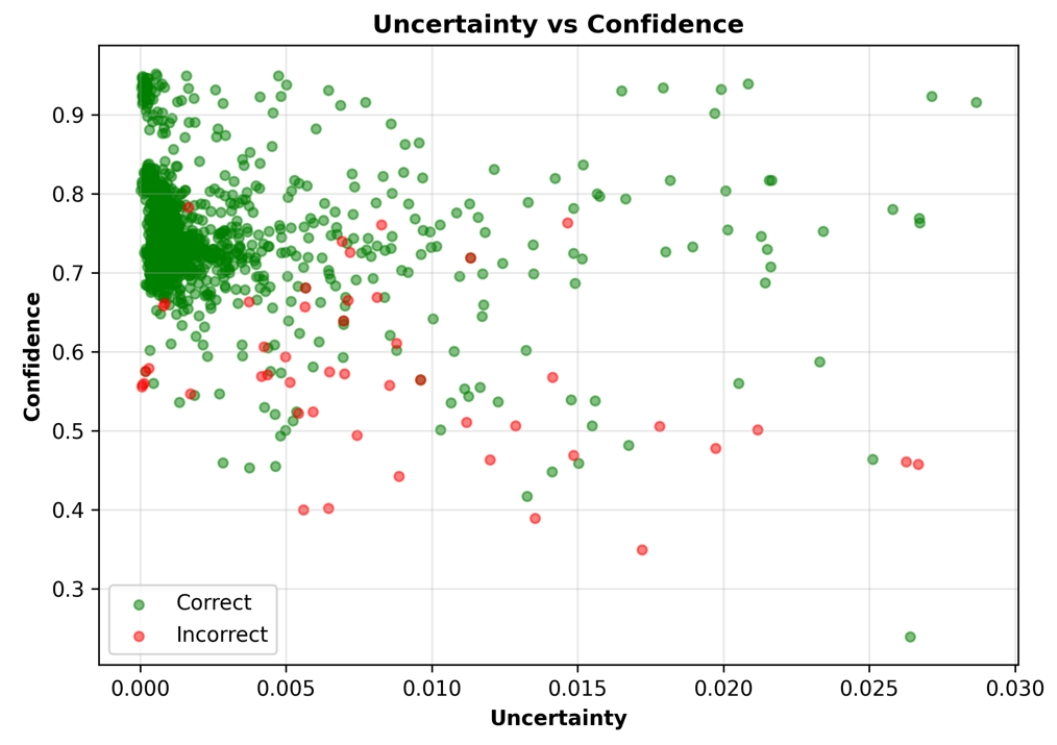}\label{fig:uncertainty_vs_confidence_all_mag_densenet}} 
    \hfill
    \subfloat[ConvNeXt U vs C]{\includegraphics[width=0.23\linewidth]{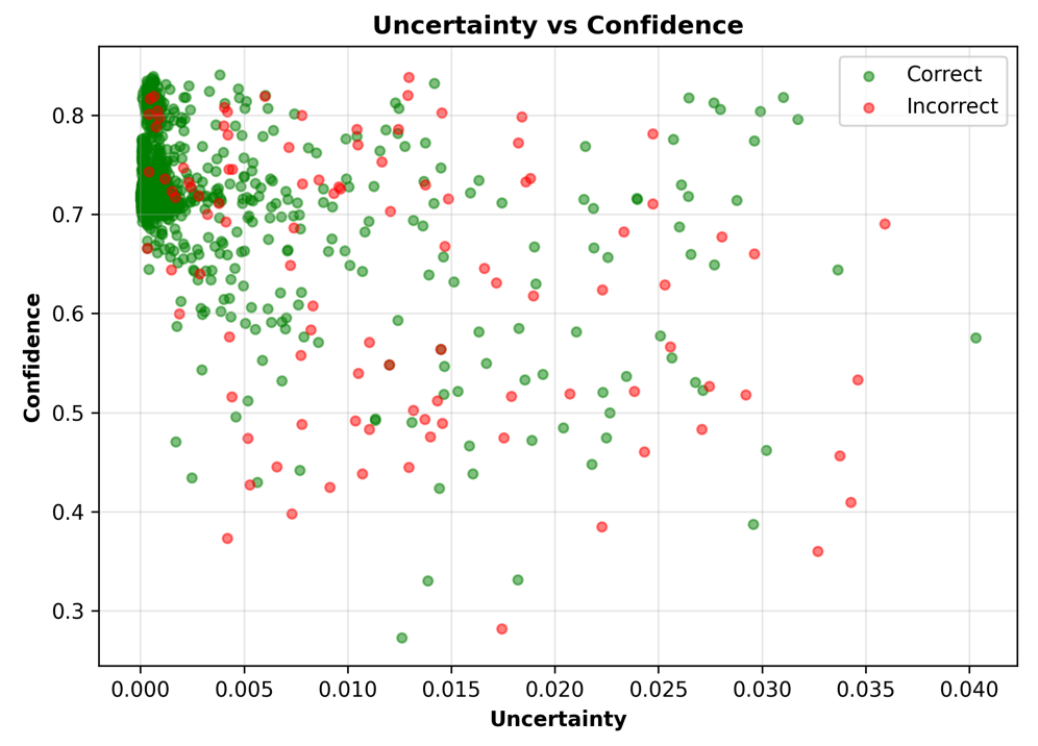}\label{fig:uncertainty_vs_confidence_all_mag_convnext}}
    
    \caption{Comprehensive visual analysis of predictive reliability. Row 1 (a-d): Cross-magnification generalization (trained at $100\times$, evaluated at $40\times$) comparing uncertainty and confidence distributions. Row 2 (e-h): Multi-magnification evaluation comparing variance and predictive confidence (U vs C) correlations for models trained and evaluated on data across all magnification levels.}
    \label{fig:combined_reliability_analysis}
\end{figure}

Visual analysis backs up this improvement in performance even more. For cross-magnification generalization, Efficient-Net-V2-S has a confidence distribution where the uncertainty for wrong predictions (red) is almost completely the same as the uncertainty for right predictions (green) (Fig. \ref{fig:uncertainity_efficientnet}, \ref{fig:confidence_efficientnet_100x_40x}). This means that the model is often wrong with confidence, which leads to outputs that are not reliable and have a lot of variation. Conversely, ConvNeXt-Tiny displays comparatively reduced variance and greater certainty in its predictions (Fig. \ref{fig:uncertainity_convNext}, \ref{fig:confidence_convnext_100x_40x}).

Also, looking at the variance and confidence distributions for multi magnification training shows that DenseNet201 strikes a good balance between accuracy and reliability(Fig. \ref{fig:uncertainty_all_mag_densenet}, \ref{fig:uncertainty_vs_confidence_all_mag_densenet}). The scatter plots, which show most of the points leaning toward the top-left corner, show that the model makes high-confidence, low-variance classifications with very few wrong guesses. There were no misclassifications above the 80\% confidence level for DenseNet201. On the other hand, ConvNeXt-Tiny had similar overall accuracy, but there were still some misclassifications even when the confidence levels were over 80\%(Fig. \ref{fig:uncertainty_all_mag_convnext}, \ref{fig:uncertainty_vs_confidence_all_mag_convnext}). These observations underscore the enhanced performance and reliability of DenseNet201 in joint-training scenarios.

\begin{table}[H]
\centering
\caption{Performance metrics of the proposed Histo-MExNet framework under the Type 3 experimental setup, which combines all magnification levels for training and tests on a stratified split.}
\label{tab:hybrid_model_performance_all_mag}
\resizebox{\textwidth}{!}{ 
\begin{tabular}{l c c c c}
\hline
\textbf{Backbone Model} & \textbf{Accuracy} & \textbf{W-Precision} & \textbf{W-Recall} & \textbf{W-F1 Score} \\ 
\hline
DenseNet201 \textcolor{green!50!black}{\large\textbf{\(\star\)}} 
& \textbf{96.97\%} & 0.9703 & 0.9697 & 0.9698\\

ConvNeXt-Tiny & 93.49\% & 0.9352 & 0.9349 & 0.9347\\

EfficientNetV2-S & 93.93\% & 0.9395 & 0.9393 & 0.939\\
\hline
\end{tabular}
} 
\end{table}

\begin{figure}[H]
    \centering
    \includegraphics[width=\linewidth]{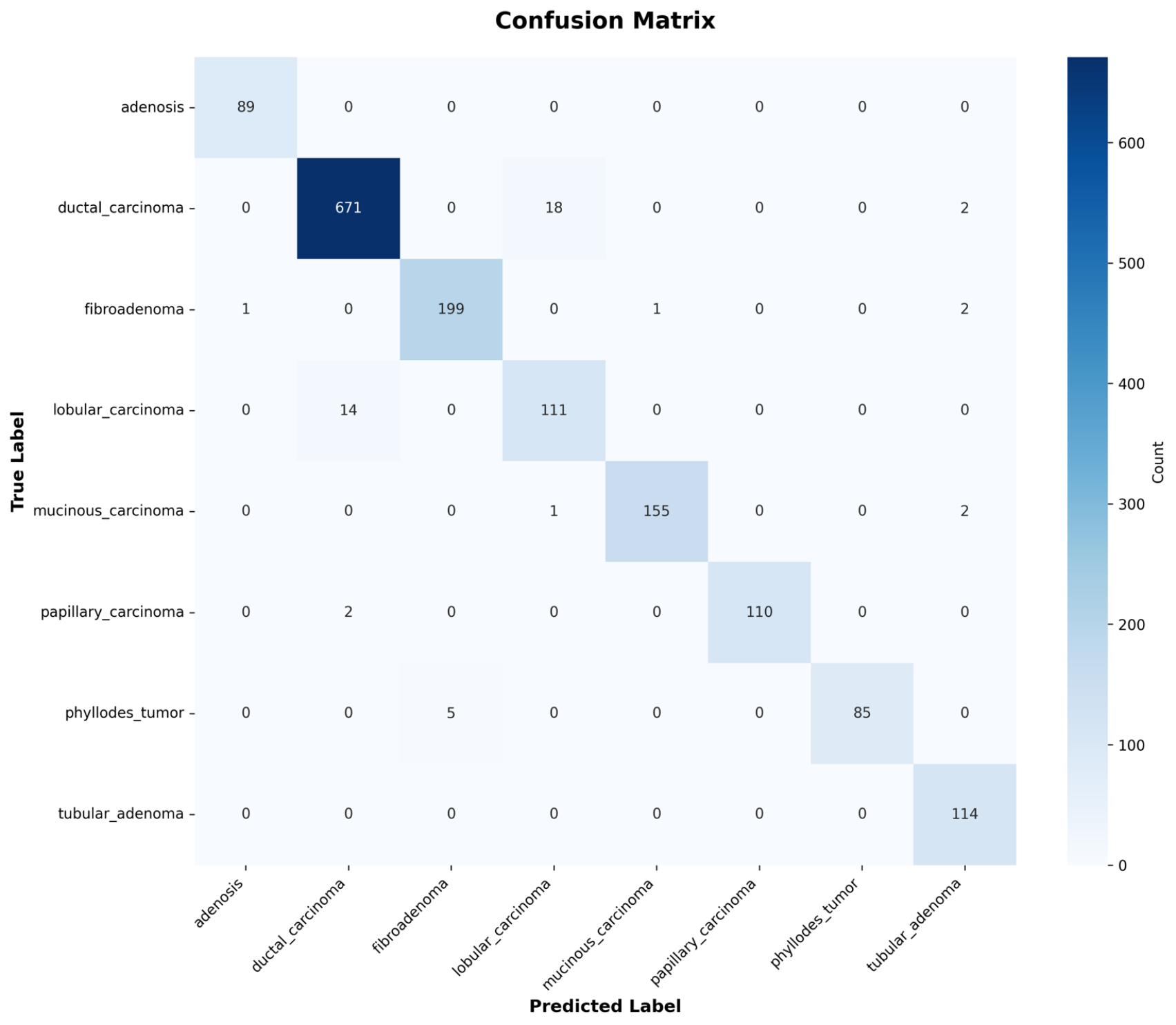}
   \caption{Confusion matrix for the DenseNet201 backbone under Type 3 experimental condition, trained on mixed data from all magnifications and assessed on a stratified test split.}

    \label{fig:confusion_matrix}
\end{figure}

The confusion matrix (Fig. ~\ref{fig:confusion_matrix}) shows that the bulk of misclassifications occur between two morphologically similar malignant sub-classes, Lobular Carcinoma (LC) and Ductal Carcinoma (DC). This trend is consistent with biological predictions, indicating that the mistakes are caused by intrinsic morphological similarities rather than random model failure.

\subsection{Interpretability and Biological Plausibility}
Although quantitative metrics demonstrate that our proposed Histo-MExNet can already makes high confidence correct prediction, quantitative metrics alone cannot rule out the "Clever Hans" phenomenon, where a model might rely on spurious background artifacts. It is equally critical to verify that the decision-making process is grounded in valid biological signals. We moved beyond the "black-box" paradigm to validate the decision-making rationale of our framework to bridge this gap between computational metrics and clinical trustworthiness. 

We had generated Occlusion Sensitivity maps and extracted three key interpretability metrics. Prediction Confidence (\%), High Sensitivity Coverage (the \% of the image area deemed critical) and Maximum Sensitivity ($S _{max}$) (the peak intensity of the most important feature).We used stratified random selection to obtain 320 representative photos, with 10 samples per class at each magnification level ($10 \times 8 \times 4 = 320$). To guarantee that the model's logic was examined under ideal operating circumstances, we limited the selection to data with a prediction confidence greater than 70\%.

\begin{table}[H]
\centering
\caption{Quantitative Analysis of Occlusion Sensitivity Features by Class and Magnifications}
\label{tab:xai_analysis}
\resizebox{\textwidth}{!}{%
\begin{tabular}{l|cc|cc|cc|cc}
\hline
\multirow{2}{*}{\textbf{Histological Subtype}} & \multicolumn{2}{c|}{\textbf{40X}} & \multicolumn{2}{c|}{\textbf{100X}} & \multicolumn{2}{c|}{\textbf{200X}} & \multicolumn{2}{c}{\textbf{400X}} \\

 & \textbf{mean} &    \textbf{max} & \textbf{mean}  &     \textbf{max} &  \textbf{mean} &     \textbf{max} &  \textbf{mean} &     \textbf{max} \\ \hline
\textit{Benign} & & & & & & & & \\
Adenosis & 1.44 & 4.6 & 0.96 & 3 & 1.68 & 4 & 1.16 & 4.8 \\

Fibroadenoma & .94 & 2.4 & 0.86 & 3.4 & .96 & 2.8 & 2.66 & 5.6 \\

Phyllodes Tumor & 0.9 & 3 & 1.12 & 3.6 & 1.2 & 4.6 & 0.58 & 1.2 \\

Tubular Adenoma & 1.26 & 3.2 & 1.78 & 3.28 & 0.98 & 3.2 & 1.68 & 5.8 

\\ \hline
\textit{Malignant} & & & & & & & & \\

Ductal Carcinoma & 1.6 & 5.8 & 0.76 & 3.6 & 3.18 & 10.8 & 2.54 & 7 \\

Lobular Carcinoma & 0.5 & 2.2 & 0.52 & 1.6 & 0.82 & 4.6 & 0.94 & 4.2 \\

Mucinous Carcinoma & 2.1 & 8.6 & 0.96 & 3.2 & 1.32 & 6.4 & 1.28 & 3.6 \\

Papillary Carcinoma & 1.34 & 3.2 & 0.95 & 3 & 2.24 & 11.8 & 1.14 & 4.6 \\

\hline
\end{tabular}%
}
\end{table}

The aggregated results of this analysis are presented in table \ref{tab:xai_analysis} reveals a distinct "scale-adaptive" attention mechanism where the model shifts its focus based on the biological features visible at different resolutions. 

The heatmaps show that the model learnt to tell the difference between benign lesions that are epithelial-dense and those that are stroma-dominant without being told to. The model concentrated heavily on the dense clusters of acini and tubules in subtypes rich in glands, particularly Adenosis and Tubular Adenoma. This visual impression corresponds with the statistics in Table ~\ref{tab:xai_analysis}, which indicates that these dense subtypes achieved higher mean sensitivity scores at $40\times$ (1.44 and 1.26, respectively).

\begin{figure}[H]
    \centering
    \includegraphics[width=0.8\linewidth]{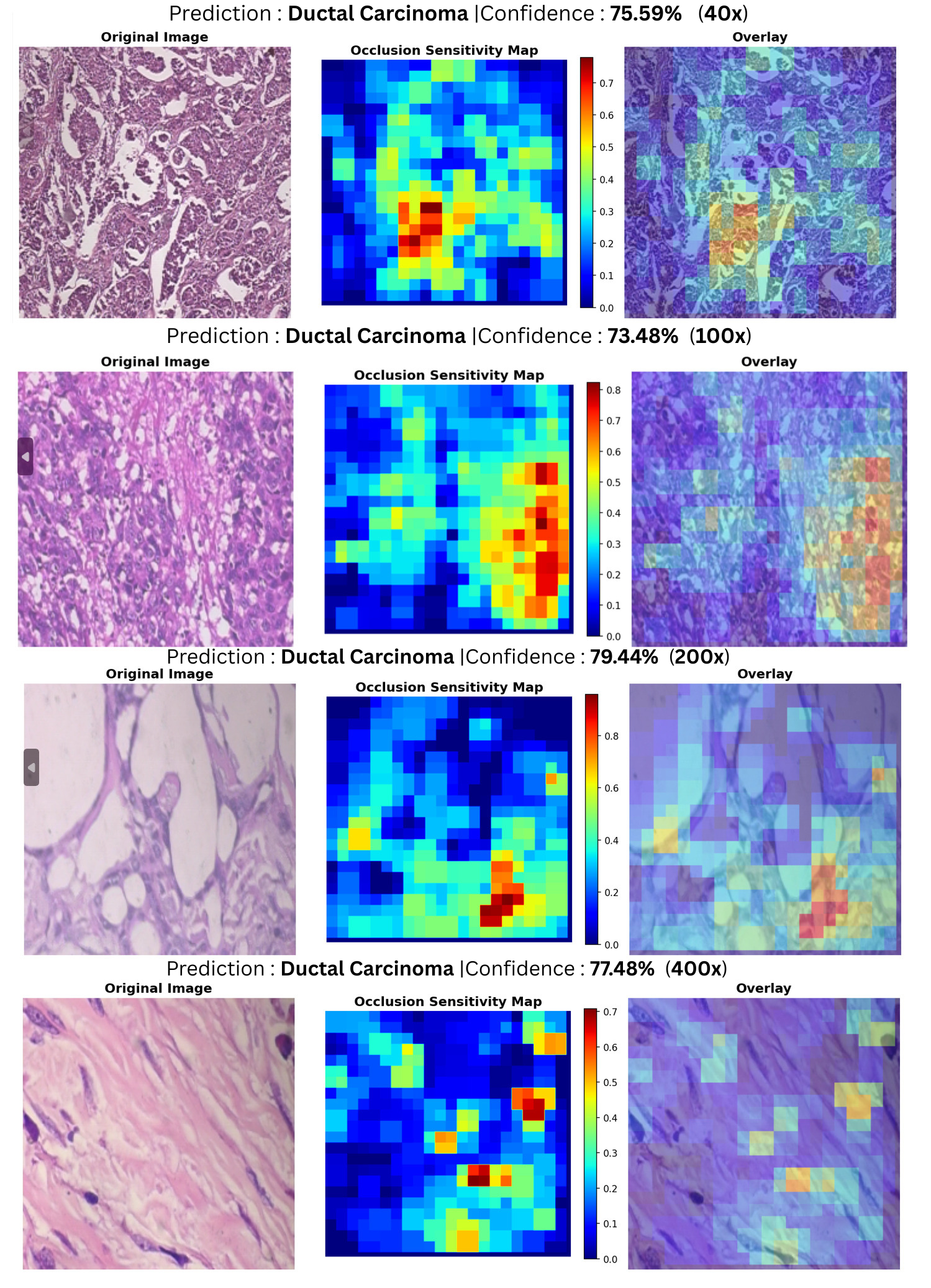}
    \caption{Occlusion sensitivity study in Ductal Carcinoma. The rows indicate magnification levels of $40\times$, $100\times$, $200\times$, and $400\times$. Columns display the input picture, occlusion sensitivity map, and overlay. The red zones emphasize the individual tissue properties that have the most influence on the model's categorization choice, with confidence ratings for each scale.}

    \label{fig:xai}
\end{figure}

On the other hand for the malignant types model identifies a clear "sweet spot" for our most difficult class Ductal Carcinoma at $200\times$, achieving a Maximum Sensitivity of 10.8 with a high Mean Sensitivity of 3.18 \ref{fig:xai}. This suggests that the model locates dense, cohesive tumor masses that are most prominent at this resolution. In contrast, Lobular Carcinoma shows consistently lower Mean Sensitivity (ranging from 0.5 to 0.94) across magnifications compared to Ductal Carcinoma. This lower more distributed attention maps to the "single-file" infiltration pattern of lobular cancer; rather than fixating on a single mass, the model aggregates weaker signals across the diffuse infiltrate.

The sensitivity maps largely aligned with correct tumor regions, a subset of heatmaps showed distributed attention without a singular focal point. We used a dual-validation method to make sure these results were correct. To assess for feature consistency, we first cross-referenced the areas of interest using a multimodal Large Vision-Language Model (LVLM). Second, and most importantly, a qualified pathologist manually reviewed the heatmaps and the original slides. The expert judgment verified that the areas that Histo-MExNet indicated, such as the mucin pools at $40\times$ and the tubular forms at $400\times$, were histologically relevant. This shows that the model's performance is based on real pathological traits and not on artifacts..

\subsection{Comparison Against State-of-the-Art for Multiclass Classification}
We tested our suggested model, the magnification-invariant ensemble, against a number of the best methods on the 8-class BreakHis dataset. Our solution, on the other hand, deals with the harder job of magnification invariance by using a single unified pipeline to categorize all magnifications. Moreover, not all studies report the same evaluation metrics; accordingly, we focus on overall accuracy as the most widely reported metric for comparison. Our proposed framework achieves state-of-the-art performance. The single, unified All-Mag (Trained on all magnification data) model attained an aggregated accuracy of 96.97\% on the hold-out test (Table \ref{tab:SOTA_ANALYSIS}). Notably, our model achieved 98.25\% accuracy on $40\times$ data, surpassing reported accuracies from other approaches. We reproduced the ES-CNN experiment ~\cite{Yusuf2025Multi} using the official model architecture and code provided in its paper within our own pipeline and obtained the accuracy reported in the comparison table. For CSDCNN ~\cite{Han2017Breast} and HF+DNN ~\cite{Joseph2022Improved}, the accuracy values included in the table correspond to those originally reported by the respective authors. In contrast, our proposed framework establishes a comprehensive end-to-end system that not only achieves State-of-the-art accuracy but also demonstrates trustworthiness by integrating variable prototypes, a physics-informed biological matrix, and a Mixture of Experts mechanism.

\begin{table}[H]
\centering
\caption{Comparative analysis of classification accuracy (\%) between existing state-of-the-art methods and the proposed Histo-MExNet across four individual magnifications ($40\times, 100\times, 200\times, 400\times$) and the combined all-magnification setting. \tnoteref{tn5}}
\label{tab:SOTA_ANALYSIS}

\setlength{\tabcolsep}{4pt}
\renewcommand{\arraystretch}{1.5}

\begin{tabularx}{\textwidth}{l *{5}{c}}
\hline
\textbf{Method} & \textbf{40X} & \textbf{100X} & \textbf{200X} & \textbf{400X} & \textbf{All-Mag} \\
\hline

ES-CNN \cite{Yusuf2025Multi} \textcolor{red!80!black}{\textbf{\(\downarrow\)}}
& 83.46\% & 79.62\% & 82.13\% & 78.3\% & 79.01\% \\
\hline

HF+DNN \cite{Joseph2022Improved}
& 90.87\% & 89.57\% & 91.58\% & 88.67\% & - \\
\hline

CSDCNN \cite{Han2017Breast}
& 89.4\% & 90.8\% & 88.6\% & 87.6\% & - \\
\hline

\textbf{Histo-MExNet} \textcolor{green!50!black}{\textbf{\(\star\)}}
& \textbf{98.25\%} & \textbf{96.88\%} & \textbf{92.8\%} & \textbf{88.46\%} & \textbf{96.97\%} \\
\hline

\end{tabularx}
\end{table}

\subsection{Ablation Studies}
We conducted an ablation study to evaluate the contribution of key components in Histo-MExNet. We had created four different version of the model where we removed one key component on each version to see the impact of that component. The corresponding configurations and results for all versions are summarized in Table~\ref{tab:ablation_table}.

\begin{table}[t]
\centering
\caption{Ablation study of Histo-MExNet evaluating the impact of key components on classification performance. The table reports Accuracy, Weighted Precision, Weighted Recall, and Weighted F1-Score for the full model and variants with specific components removed, including ensemble learning (A1), hybrid loss (A2), attention module (A3), and prototype module (A4). This analysis highlights the individual contribution of each module to the overall model effectiveness.}

\label{tab:ablation_table}

\renewcommand{\arraystretch}{1.35}

\begin{tabularx}{\textwidth}{l *{4}{c}}
\hline
\textbf{Variant} & \textbf{Accuracy (\%)} & \textbf{W-Precision} & \textbf{W-Recall} & \textbf{W-F1} \\
\hline
\textbf{Histo-MExNet} & \textbf{96.97} & \textbf{0.9703} & \textbf{0.9697} & \textbf{0.9698} \\
A1 & 93.05 & 0.9303 & 0.9305 & 0.9296 \\
A2 & 90.49 & 0.9051 & 0.9049 & 0.9048 \\
A3 & 91.81 & 0.9175 & 0.9181 & 0.9176 \\
A4 & 92.62 & 0.9259 & 0.9262 & 0.9259 \\
\end{tabularx}

\vspace{8pt}

\noindent\footnotesize
\textbf{Descriptions of Ablation Variants}\\[4pt]

\begin{tabularx}{\textwidth}{p{0.40\textwidth} X}
\textbf{Histo-MExNet (Proposed)} &
Complete model with ensemble learning, hybrid loss formulation, attention mechanism, and prototype module. \\[6pt]

\textbf{A1 – No Ensemble} &
Uses only the single best-performing fold instead of 5-fold ensemble aggregation. \\[6pt]

\textbf{A2 – Cross-Entropy Only} &
Optimized exclusively with standard cross-entropy, removing hybrid loss components. \\[6pt]

\textbf{A3 – No Attention Module} &
Removes spatial–channel attention, relying solely on the base feature extractor. \\[6pt]

\textbf{A4 – No Prototype Module} &
Excludes prototype-based representation learning and uses conventional feature embeddings. \\
\hline
\end{tabularx}

\end{table}

We began the ablation study by evaluating the contribution of the ensemble strategy after cross validation. Instead of the full 5-fold weighted aggregation we cherry-picked only single best performing fold in variant $A1$. This modification directly caused to a marked drop in performance (\textbf{F1: 0.9296 vs. 0.9698}). It means that the ensemble helps reduce fold-specific variability and improves robustness specifically in challenging cases. After this we evaluated the role of the hybrid loss in variant $A2$ by substituting it with standard cross-entropy loss. It results in the lowest F1-score among all variants (\textbf{0.9048}). This means that hybrid loss plays an important role in histopathology-informed guidance that strengthens class separation.
The result after removing the attention mechanism in $A3$ revealed a moderate decline in performance (\textbf{F1: 0.9176}). It means that there is an importance of spatial and channel-wise refinement in drawing diagnostically relevant structures that help our backbone. As the last ablation variant $A4$ we eliminated the prototype module, that results in a moderate reduction in performance (\textbf{F1: 0.9259}). That means prototype-based representations help to improve discriminability and more informative decision boundaries in regions with overlapping textures. In conclusion, the ablation results show that each component contributes in a distinct but complementary way to the effectiveness of our Histo-MExNet framework.

\section{Discussion}
\label{sec:discussion}

\subsection{Overcoming Domain Shift and Data Challenges}
The baseline experiments (Table \ref{tab:performance_baseline_models}) revealed a significant limitation of standard Convolutional Neural Networks (CNNs) in the context of real-world histopathological data: they exhibit difficulties in generalizing under conditions of pronounced class imbalance and substantial morphological variability. Histo-MExNet addresses this problem not by inflating minority classes, but by limiting the hypothesis space via physics-informed regularization and prototype-driven representation learning. The biologically grounded loss terms contain explicit priors on tissue morphology, directing the optimization process towards stable features under real-world morphological variability.

Our cross-magnification analysis (Table \ref{tab:hybrid_model_performance_merged}) confirms that magnification changes constitute a substantive covariate shift. At lower magnifications ($40\times$), the visual signal is dominated by global structural arrangements, such as the organization of lobules and ducts. In contrast, higher magnifications ($400\times$) are dominated by cellular-level textures, such as nuclear atypia and mitotic figures. Our results indicate that joint training (Stage 3) is the definitive answer to this domain shift. By exposing the Histo-MExNet to a stratified mix of all magnifications, the model is forced to learn scale-invariant feature representations, effectively bridging the semantic gap between global tissue architecture and local cellular morphology.

\subsection{Clinical Trustworthiness, Limitations, and Future Work}
The dependability of prediction is as important as its accuracy in medical diagnosis. Our comparative study (Fig. ~\ref{fig:combined_reliability_analysis}) reveals that some conventional models, such as Efficient NetV2-S, can show harmful overconfidence. In clinical practice, "confident errors" are particularly problematic because they can result in misdiagnoses without any signal that additional review is needed. Histo-MExNet helps to mitigate this risk by achieving better calibration through the combined use of Bayesian Monte Carlo Dropout and a multi-expert ensemble. Empirically, the model did not produce misclassifications when the predicted confidence was close to 80\%, suggesting a practical operating point: predictions with confidence above this threshold can be accepted with high trust, while those below it should be flagged for pathologist review.

Through the use of prototype learning, the framework improves interpretability by allowing query samples to be directly compared with learned class-specific representations, rather than relying solely on opaque black-box predictions. Despite its strengths, the proposed methodology has several limitations. The most prominent is the increased computational overhead introduced by the ensemble architecture. In practice, this reflects an explicit trade-off: the framework favors clinical robustness at the cost of slower inference, highlighting the need for further optimization before deployment in resource-limited environments.

Demonstrating broader clinical generalizability will therefore require additional validation on multicenter datasets that capture variability in scanners, staining protocols, and sample preparation. A further challenge is the reliable discrimination of physiologically similar subtypes, such as Lobular and Ductal Carcinoma. As indicated by the confusion matrices (Fig. \ref{fig:confusion_matrix}), their substantial morphological overlap remains a key source of misclassification. To address this, future studies should explore multi-modal approaches that integrate histopathological features with genomic or clinical information, and aim to extend the present patch-based classification framework to Whole Slide Image (WSI) analysis.

\section{Conclusion}
\label{sec:conclusion}

This study presents Histo-MExNet to address magnification-dependent domain shifts and morphological heterogeneity. The model integrates a multi-expert ensemble of convolutional backbones with physics-informed regularization. Evaluated on the BreaKHis dataset, Histo-MExNet attains state-of-the-art performance, achieving 96.97\% overall accuracy under multi-magnification training. Beyond predictive performance, the framework explicitly incorporates interpretability and robustness. A prototype learning module anchors classification decisions to representative exemplar cases, improving transparency of decision boundaries. Sensitivity Occlusion analysis demonstrates that the model consistently attends to diagnostically relevant cellular structures rather than background artifacts. Furthermore, uncertainty estimation via Monte Carlo Dropout enhances reliability by identifying ambiguous and out-of-distribution samples. Ablation studies confirm that embedding biological constraints mitigates overfitting to staining-related artifacts and improves generalization to unseen magnification levels. Overall, these findings demonstrate that integrating expert diversity, prototype guidance, uncertainty quantification, and domain-informed regularization enables a balanced framework that maintains strong predictive performance while improving robustness, interpretability, and clinical applicability.





\vspace{4mm}

\textbf{CRediT authorship contribution statement}\\
\textbf{Enam Ahmed Taufik:} Conceptualization, Formal analysis, Methodology, Writing – original draft, Writing – review and editing, \textbf{Md Ahasanul Arafath:} Conceptualization, Formal analysis, Methodology, Writing – original draft, Writing – review and editing, \textbf{Abhijit Kumar Ghosh:} Formal analysis, Writing – original draft, Writing – review and editing, \textbf{ Md.
Tanzim Reza:} Writing – review and editing, Supervision,  \textbf{Md Ashad Alam:} Writing – review and editing, Supervision.

\vspace{4mm}

\textbf{Data and code availability}\\
The BreaKHis dataset used in this study is publicly available. The full dataset can be accessed through the official repository: \href{https://web.inf.ufpr.br/vri/databases/breast-cancer-histopathological-database-breakhis/} {Breakhis}. A curated version is also available on Mendeley Data. All data were used in compliance with the dataset’s licensing terms.


\vspace{4mm}

\textbf{Declaration of competing interest}\\
The authors declare that they have no known competing financial interests or personal relationships that could have influenced the work reported in this paper.

\end{document}